\documentclass[letterpaper, 10 pt, conference]{ieeeconf}
\IEEEoverridecommandlockouts
\usepackage{preamble}

\title{\LARGE \bf
From Language to Action:\\Can LLM-Based Agents Be Used for Embodied Robot Cognition?
}

\author{Shinas Shaji$^{*,\dagger,\mathsection}$, Fabian Huppertz$^{*,\dagger}$, Alex Mitrevski$^{\ddagger}$, and Sebastian Houben$^{\dagger,\mathsection}$
\thanks{This work was supported by the b-it foundation} %
\thanks{$^{\dagger}$Institute of AI and Autonomous Systems (A$^2$S), Hochschule Bonn-Rhein-Sieg, Sankt Augustin, Germany,
        {\tt\scriptsize <shinas.shaji, fabian.huppertz>@smail.inf.h-brs.de, sebastian.houben@h-brs.de}} %
\thanks{$^{\ddagger}$Division for Systems and Control, Chalmers University of Technology, 41258 Gothenburg, Sweden, {\tt\scriptsize alemitr@chalmers.se}}\newline
\thanks{$^{\ddagger}$Fraunhofer Institute for Intelligent Analysis and Information Systems, Sankt Augustin, Germany, {\tt\scriptsize <shinas.shaji, sebastian.houben>@iais.fraunhofer.de}}\newline
\thanks{$^{*}$Corresponding author} %
}

\begin{document}

\maketitle
\thispagestyle{empty}
\pagestyle{empty}


\begin{abstract}
    In order to flexibly act in an everyday environment, a robotic agent needs a variety of cognitive capabilities that enable it to reason about plans and perform execution recovery.
    Large language models (LLMs) have been shown to demonstrate emergent cognitive aspects, such as reasoning and language understanding; however, the ability to control embodied robotic agents requires reliably bridging high-level language to low-level functionalities for perception and control.
    In this paper, we investigate the extent to which an LLM can serve as a core component for planning and execution reasoning in a cognitive robot architecture.
    For this purpose, we propose a cognitive architecture in which an agentic LLM serves as the core component for planning and reasoning, while components for working and episodic memories support learning from experience and adaptation.
    An instance of the architecture is then used to control a mobile manipulator in a simulated household environment, where environment interaction is done through a set of high-level tools for perception, reasoning, navigation, grasping, and placement, all of which are made available to the LLM-based agent.
    We evaluate our proposed system on two household tasks (object placement and object swapping), which evaluate the agent's reasoning, planning, and memory utilisation.
    The results demonstrate that the LLM-driven agent can complete structured tasks and exhibits emergent adaptation and memory-guided planning, but also reveal significant limitations, such as hallucinations about the task success and poor instruction following by refusing to acknowledge and complete sequential tasks.
    These findings highlight both the potential and challenges of employing LLMs as embodied cognitive controllers for autonomous robots.
\end{abstract}


    \section{INTRODUCTION}
    \label{sec:introduction}

    Robotic systems for everyday environments need various high-level abilities, such as complex task planning, execution monitoring, and failure recovery, so that they can flexibly perform a wide variety of tasks.
    Significant work has been done on systems with such broad capabilities in the context of cognitive agent architectures \cite{kotseruba2018,laird2012,franklin2014,beetz2018}, which equip systems with an ability to plan actions, perceive their environment, act accordingly, reason about the world, learn, and adapt \cite{corobook}.
    Traditional cognitive architectures, however, usually require a large number of dedicated components that handle concrete aspects of cognition, or they depend on large amounts of engineered prior knowledge to work effectively.
    This has generally limited the widespread adoption of such architectures on most robotics platforms.

    In the last few years, large language models (LLMs) and vision-language models (VLMs) have been shown to exhibit notable abilities in reasoning, language understanding, and task execution \cite{liang2023,huang2023,wangj2024,mower2024,wangg2024,tang2025,openx2024}.
    Importantly, the robot's (and by extension, the LLM agent's) planning and decision-making capabilities take on a more grounded role, reasoning in natural language about the physical consequences of movements and manipulative actions.
    This opens new directions for exploring embodied cognition and could inform future developments in cognitive robotics, assistive AI, and multi-modal agent systems.
    However, key challenges arise in bridging high-level natural language reasoning with low-level embodied actions.
    The agent must interpret abstract task descriptions and translate these into sequences of precise robotic actions such as navigation, grasping, and placement.
    This translation process is complicated by the inherent ambiguity in natural language instructions, leading to misinterpretations where agents may attempt to manipulate inappropriate objects.

\begin{figure*}[t]
        \centering
        \includegraphics[width=0.85\linewidth]{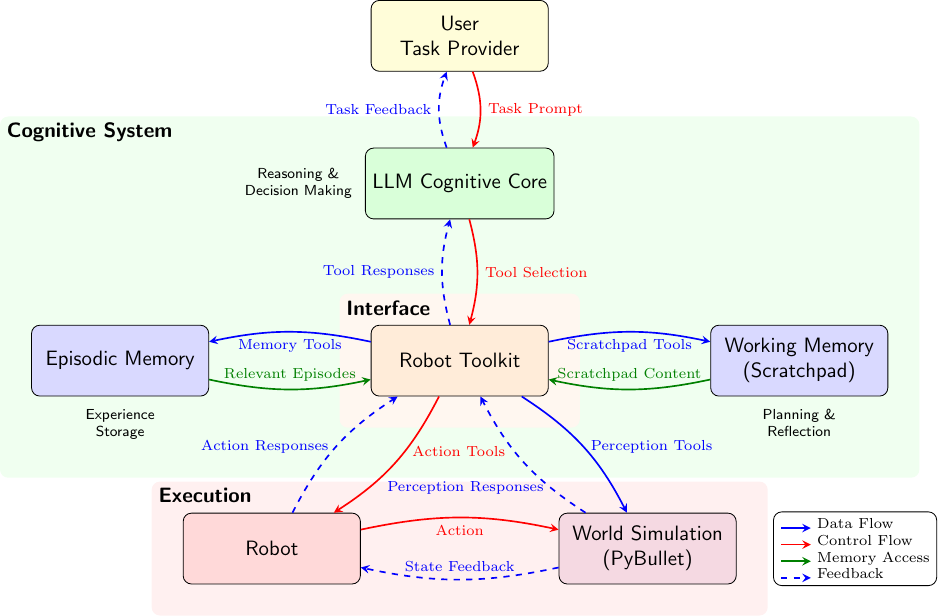}
        \caption{Overview of our proposed system}
        \label{fig:overview}
    \end{figure*}

    This paper investigates whether LLMs (used in a zero-shot manner) can reasonably function as the core of a cognitive architecture within a simulated 3D robotic setting, where the agent physically manipulates objects and executes tasks in the real world.
    By assigning the agent physical tasks in a simple household environment, we probe how LLMs reason about spatial layouts, interpret user instructions, and adapt plans based on feedback; through this, we aim to investigate whether such systems can perceive, plan, act, and adapt in grounded, interactive settings.
    Additionally, we explore how episodic memory can enhance task performance through experience accumulation.
    Our system is illustrated in Fig. \ref{fig:overview}.

    We concretely propose a cognitive agent architecture that embeds an LLM as its central reasoning engine and situates it in a 3D simulation of a simplified household environment.
    Here, the robot is embodied as a mobile manipulator with a 7-DOF arm mounted on an omnidirectional base, enabling navigation and object manipulation.
    The environment features a kitchen and living room layout, in which the robot must carry out tasks such as pick-and-place or sorting objects between designated areas.
    The agent receives observations from the environment primarily as natural language descriptions, detailing the positions and types of objects and entities.
    Memory is organised into two main tiers: \emph{working memory} in the form of a prompt for immediate task information, and semantically searchable \emph{episodic logs} for synthesised summaries of past actions.
    The LLM is prompted with chain-of-thought prompting \cite{wei2022} to reason and form plans accounting for objects, entities, and anticipated outcomes.
    A set of high-level tool functions were implemented to expose perception, navigation, grasping, and placement as callable actions for a function-calling LLM \cite{shen2024}.
    Object placement and shelf reordering were chosen to test the agent's ability to execute complex high level tasks, utilize its memory system, and adapt to constraints such as occupied spaces.

    The contributions of this work include:
    (i) the \emph{integration of an agentic LLM into a cognitive architecture} with memory and tool interfaces, allowing a robot to interact with its environment through perception, movement, and object manipulation, and
    (ii) \emph{empirical evaluation of our approach in a simulated 3D environment on two household tasks}, which provides insights into the strengths and limitations of current LLMs when applied to embodied decision-making.
    The implementation of our method is publicly available.\footnote{\url{https://github.com/ShinasShaji/llm-robot-cognition}}

    \section{RELATED WORK}
    \label{sec:related-work}

    \paragraph{Cognitive architectures} Cognitive architectures provide comprehensive frameworks for developing intelligent systems with human-like cognitive capabilities \cite{kotseruba2018,corobook}.
    Notable architectures include the SOAR architecture \cite{laird2012}, which integrates symbolic reasoning with perceptual and motor systems for cognitive robotics; the architecture features mental imagery, episodic and semantic memory, reinforcement learning, and continuous model learning capabilities that enable robust robotic control in complex environments.
    Similarly, ACT-R \cite{ritter2019} offers a hybrid symbolic-subsymbolic framework for modelling human cognition, with detailed memory models, perceptual-motor integration, and strong empirical grounding in cognitive psychology.
    A notable robotics-specific architecture is KnowRob 2.0 \cite{beetz2018}, which combines rich knowledge (in the form of ontologies) with interfaces for logical querying and reasoning, narrative-enabled episodic memories, components for metacognition, and a simulation for physics-based reasoning.
    Such traditional architectures typically require extensive engineered knowledge and a large number of dedicated components, which makes it challenging to adapt the approaches to diverse robotic platforms.
    Unlike traditional symbolically grounded approaches, our system leverages LLMs' natural language capabilities for high-level reasoning and flexible task adaptation, while maintaining the embodied, interactive aspects crucial for robotic cognition.

    \paragraph{LLM-based embodied AI} Recent advances in LLMs have introduced new possibilities for developing embodied AI systems \cite{liang2023,huang2023,wangj2024}; these correspond to the emergent cognition paradigm \cite{corobook}.
    Concretely, LLM-based agents can, in principle, interpret natural language instructions, reason about spatial relationships, and generate complex plans without extensive domain-specific engineering, leveraging their pre-trained knowledge and reasoning capabilities \cite{huang2023a}.
    However, these approaches often face challenges in grounding abstract reasoning to concrete robotic actions, managing long-term task execution, and maintaining consistency in multi-step operations \cite{backlund2025}.

    Complementary work in embodied AI has explored reinforcement learning agents within simulation frameworks such as MineDojo \cite{fan2022}, demonstrating robust performance in constrained environments. 
    Generative models have shown promise for creating social agents that simulate believable human behaviour through architectures that maintain memory streams and enable reflection \cite{park2023}, while open-ended embodied agents such as Voyager \cite{wangg2024} demonstrate lifelong learning capabilities by incrementally building skill libraries in complex environments.
    However, these approaches often face challenges in grounding abstract task commands to concrete robotic actions and exhibiting agency.

    In \cite{manling2024}, a comparison of various LLMs is performed for modular planning abilities such as subgoal decomposition and action sequencing, relying on a traditional logical representation of the world and goal. 
    On the other hand, in \cite{monwilliams2025}, an LLM is used in conjunction with a pre-defined knowledge base of code snippets, where the LLM is used to generate executable code to perform sub-actions, similar to \cite{liang2023}.
    Unlike \cite{manling2024}, we rely on natural language formulations, using the emergent capabilities of the LLM in understanding tasks and sequencing actions through agentic tool calling. 
    In contrast to \cite{monwilliams2025}, we evaluate multiple models, and rely on tool calling (instead of generating executable code) to guarantee consistent behaviors composed from well-defined actions.
    Our agent also accumulates memory over repeated executions, which is not the case in \cite{manling2024,monwilliams2025}.

    \paragraph{Robot foundation models} Generalist vision-language and vision-language-action policies, such as Octo \cite{octo2024}, $\pi_0$ \cite{black2024}, RT-X \cite{openx2024}, and OpenVLA \cite{kim2024} map multimodal observations to diverse robot actions and tasks.
    These are reactive execution models, which do not have the capability to reason and deliberate, and hence are on the level of lower-level skill models.
    However, they could eventually be integrated into the execution layer of our architecture; this is an interesting direction that could be explored in future work.

    \paragraph{Agentic LLMs} Recent research on LLM agency distinguishes between \textit{AI agents} (task-specific automation systems) and \textit{agentic AI} (marked by dynamic task decomposition, persistent memory and even multi-agent collaboration) \cite{sapkota2025,schneider2025}.
    The concept of \textit{agency}, or the capacity to proactively shape events through one's actions, has emerged as a critical aspect of LLM behaviour, and represents a progression from isolated reactive agents to systems capable of adaptive, system-level intelligence \cite{sharma2024}.

    Our work bridges aspects of traditional cognitive architectures such as episodic and working memories, with agentic planning and reasoning.
    We particularly investigate whether agentic LLMs can serve as the core of a cognitive architecture for embodied robots, similar to how traditional cognitive architectures provide integrated memory structures and reasoning capabilities.

    \section{METHODOLOGY}
    \label{sec:methodology}

    In this paper, our goal is to investigate whether an LLM-based system is able to meaningfully perform high-level cognitive processes that a robot needs to execute a variety of tasks in everyday environments.
    To achieve this, we investigate an agent architecture that enables a robot to
    (i) \emph{generate task plans} as sequences of actions to satisfy natural language task instructions,
    (ii) \emph{preserve episodic memories} to simplify the planning process in later executions,
    (iii) perform \emph{online execution recovery} and replanning in case of detected failures.
    The processing of information in our system is completely done by an LLM, which receives (or requests) all details about the environment in natural language format and relies on reasoning performed by the LLM.
    Task plans are generated as sequences of natural language actions, which the LLM can execute by calling from a set of tools (such as \texttt{Look-Around} or \texttt{Move-To}) that are available to the robot.
    Here, tools can be thought of as actions in a classical planning sense, which allow us to constrain the robot's execution to what it can actually perform.
    In the rest of this section, we describe the test environment and the different components of our architecture in more detail.

    \subsection{Simulated World}

\begin{figure}[t]
        \begin{subfigure}[t]{0.475\linewidth}
            \centering
            \includegraphics[width=\linewidth]{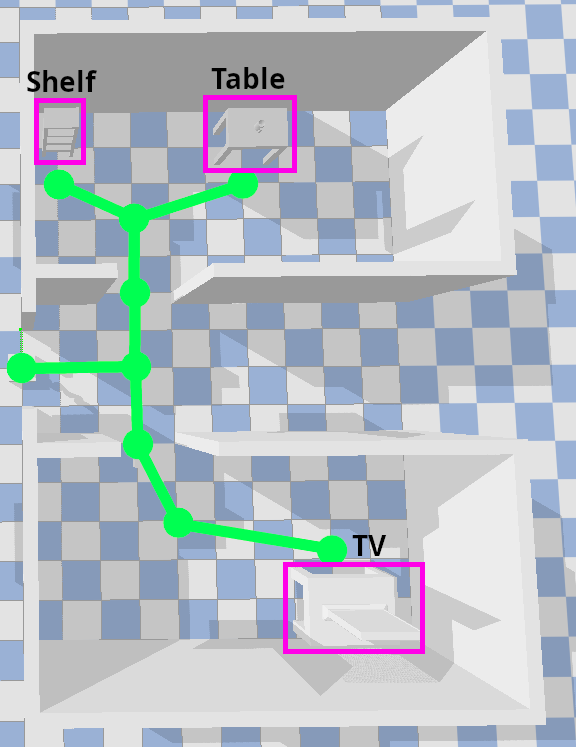}
            \caption{The simulated household environment and its semantic map representation, consisting of a kitchen, a living room, and a connecting hallway.}
            \label{fig:environment}
        \end{subfigure}
        \hspace{0.025\linewidth}
        \begin{subfigure}[t]{0.475\linewidth}
            \centering
            \includegraphics[width=\linewidth]{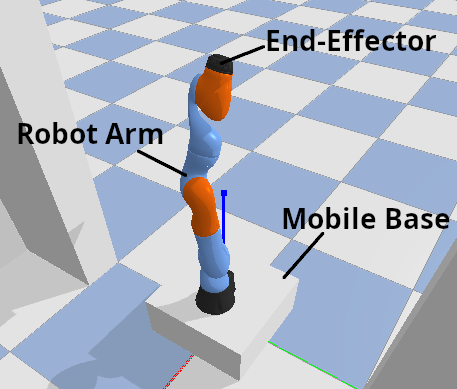}
            \caption{The simulated mobile manipulator consisting of a square omnidirectional base and a 7-DOF arm.}
            \label{fig:agent}
        \end{subfigure}
        \caption{Illustration of the simulated robot and test environment. Green dots are semantic locations that the agent can navigate to via the connected lines.}
        \label{fig:simulation-environment}
    \end{figure}

    To ground the investigation of the proposed architecture in a concrete scenario, we use a simulated robot and a 3D environment that represents a simplified two-room household, comprising a kitchen and a living room that are connected by a hallway.
    The kitchen contains a three-layer shelf and a table, both capable of supporting objects, while the living room contains a television placed on a table.
    Figure \ref{fig:environment} shows the environment and a semantic map used for navigation.

    The agent and its environment are simulated using the PyBullet physics engine.\footnote{\url{https://pybullet.org/}}
    As PyBullet is a Python module, its functionality can be directly integrated into any Python script; this allows precise control over the simulation and its interaction with external tool calls.

    The agent's body is modeled as a mobile manipulator, consisting of a square omnidirectional base and a 7-DOF robotic arm, which we derive from the internal PyBullet KUKA LBR iiwa model.
    To increase manoeuvrability, the rotational joint limits of the arm's joint that connects the base to the arm are disabled, enabling full 360$^o$ reach around the base.
    A fully articulated gripper is not implemented; instead, as mentioned before, grasping and placing were simplified by attaching objects directly to the end-effector when it is within a predefined proximity threshold, and detaching them at the desired placement location.
    Figure \ref{fig:agent} depicts the robot.

    \subsection{Tools for World Interaction}

    To interact with the environment and its objects, the agent is equipped with several tool functions, each exposed through a tool-calling interface with text-based status responses.
    The implemented tools are summarized below:
    \begin{description}
        \item[\texttt{Look-Around}] \hspace{1.35cm} Returns the agent's location on a semantic map of the environment, as well as a list of objects detected in proximity to the agent and where these objects are placed.\footnote{As perception as such is not the main focus of this paper, we do not process visual data in this work, but perception is performed based on read out information from the simulated environment.}
        \item[\texttt{Move-To}] \hspace{0.5cm} Executes navigation to a goal location on the semantic map.
        Path planning is performed using the A* algorithm \cite{hart1968}; if a valid path is found, the agent follows it until the goal is reached.
        If no path is available, the tool reports a failure.
        \item[\texttt{Grab}] Moves the robot arm towards a specified target object.
        If the end-effector reaches within a proximity threshold, the object is attached to the arm, and the tool reports success.\footnote{Due to limitations of the simulator, objects are considered grasped if the robot moves close to the object; on a real-robot system, a suitable grasp planner needs to be used in the implementation of this tool.}
        Failure is reported if the target object does not exist, is out of reach, or if the agent is already holding an object.
        \item[\texttt{Place}] \hspace{0.05cm} Allows the agent to release the currently held object at a specified location.
        Placement succeeds if the end-effector reaches the designated location, the location is unoccupied, and the agent is holding an object.
        Otherwise, the tool reports a failure, specifying the violated condition.
        \item[\texttt{Add-To-Scratchpad}] \hspace{2.5cm} The agent can write down thoughts and plans for reasoning and reflection.
        This allows the agent to plan, reason, and organise its thoughts before taking action.
        An example of scratchpad use is given in lines 13--21 of Listing \ref{lst:llm_output}.
        \item[\texttt{View-Scratchpad}] \hspace{2.1cm} The agent can view the current contents of its scratchpad, which contains previous thoughts and plans joined as paragraphs.
        \item[\texttt{End-Task}] \hspace{0.8cm} The agent ends the current task with a detailed task report, including (i) a description of the task prompt, (ii) the task status (whether it believes the task succeeded or failed), and (iii) a summary of actions taken during the task execution.
        The task report is saved to the episodic memory and is made semantically searchable through the task description.
        An example of a task report can be seen in lines 37--40 of Listing \ref{lst:llm_output}.
        \item[\texttt{Search-Memory}] \hspace{1.75cm} The agent queries its episodic memory using a description of the current task to gain information from prior executions of tasks.
        The tool returns relevant past experiences that might help with the current task, where the relevance is determined based on semantic similarity between the query and the memorised descriptions.
    \end{description}
    It is important to mention that, unlike classical planning systems, we do not perform any explicit knowledge base updates upon executing the world interaction tools (such as \texttt{Grab} and \texttt{Place}); instead, the evolution of the world is implicitly tracked by the LLM agent itself.
    Lines 23--29 of Listing \ref{lst:llm_output} illustrate a call to the \texttt{Grab} tool and its response.

    \subsection{LLM-Driven Cognitive Processes in the Architecture}
    
    The robot's cognitive architecture, illustrated in Fig. \ref{fig:overview}, is built around a large language model (LLM) that serves as the central decision-making component.
    This LLM-based approach allows the robot to process natural language instructions, reason about its environment and own plans, and select appropriate actions through a tool-calling interface.
    At the core of the robot's cognition system is its memory architecture, which consists of two distinct components that mirror cognitive processes \cite{corobook} (episodic memory and working memory), as well as its perception and action execution components.

    \paragraph{Episodic memory}
    The episodic memory is implemented using ChromaDB\footnote{\url{https://www.trychroma.com}}, a vector database that stores the robot's past experiences and makes it searchable using text embeddings.
    This allows the robot to perform similarity searches through its previous interactions, retrieving relevant experiences that can inform current decision-making processes.
    When the robot encounters a new task, it can query this episodic memory to find similar past scenarios, and subsequently adapt successful strategies or prevent unsuccessful execution attempts from those experiences.

    \paragraph{Working memory}
    The working memory of the system is represented by a combination of the LLM's context and information stored through a scratchpad mechanism, which dynamically incorporates environmental information, task descriptions, and the robot's internal reasoning during planning.
    Concretely, the current perceptual data from the environment and results of actions and tool calls are stored in the LLM's context, which the LLM agent can then write to the scratchpad; this information ensures that the robot's decisions are based on the most recent state of the world.
    Here, the objective of the scratchpad is twofold: it enables storing information that is no longer included in the current context, and it can facilitate the agent's reasoning by providing a dedicated workspace to plan and reason on the available information.

    \paragraph{Perception and action execution}
    The perception and action execution module consists of the functions in the aforementioned toolkit, which translate high-level decisions into concrete robotic actions.
    Specifically, the perception component gathers information about the robot's current location, nearby objects, and environmental layout, which is then integrated into the working memory to inform decision-making; the action components perform the respective actions on the robot and monitor the execution status.
    Here, the status is obtained based on implemented monitors in the respective tools, which verify aspects such as whether an object of interest is present or whether the robot is already holding an object.

    This cognitive architecture enables the robot to combine long-term learning from past experiences with immediate environmental awareness, allowing for adaptive and context-aware behaviour.
    In addition, as a result of the execution monitoring, the agent can identify failures during plan execution and attempt to modify its plan to ensure successful completion.
    Here, it should be mentioned that replanning is not an aspect that is explicitly enforced, but it rather emerges during the LLM reasoning process itself.

    An example output showing the task planning, reasoning, and execution process is given in Listing \ref{lst:llm_output} below, for the task of putting items away into a cupboard (with memory of previous executions).
    Here, we can see that the model queries the memory for relevant experiences, uses the scratchpad to reason over the current state of the environment and plan actions, and calls tools available to it to execute the plan. In this case, the model thinks the task was successfully completed, which is corroborated by a check of the state of the world.

    \lstset{basicstyle=\ttfamily\tiny,
        backgroundcolor=\color{gray!10!white},
        frame=lines,
        breaklines=true,
        showstringspaces=false,
        numbers=left,
        numbersep=3pt,
        xleftmargin=8pt,
        caption={\centering Example planning and action log output for Claude 4 Sonnet. The output is shortened for brevity.},
        label={lst:llm_output},
        literate=
            {\[SYSTEM\]}{{\textcolor{orange!80!black}{[SYSTEM]}}}8
            {\[TOOL\]}{{\textcolor{blue!80!black}{[TOOL]}}}6
            {\[TOOL-LOG\]}{{\textcolor{magenta!80!black}{[TOOL-LOG]}}}{10}
            {\[TOOL-RESPONSE\]}{{\textcolor{green!40!black}{[TOOL-RESPONSE]}}}{15},
        procnamekeys={def class}}
\begin{lstlisting}
[SYSTEM] Executing prompt: "Put away all the items into the shelf."
---
[SYSTEM] Robot is at location: Front Door
---
[TOOL] Robot is querying memories with the query: put away items shelf organize
[TOOL-RESPONSE] The query returned 2 results of previous experiences.
---
[TOOL] The agent wrote to the scratchpad: "Looking at my memory, I have successfully completed this exact task before! ...
I need to check the current state of the environment..."
---
[TOOL] Robot looked around...
---
[TOOL] The agent wrote to the scratchpad: "Perfect! The environment is exactly as I remember from my previous successful completion...
Based on my previous experience, I'll use the same efficient approach:
1. Grab cube from Front Door (current location)
2. Go to Kitchen Shelf and place cube on top position
3. Go to Living Room and grab box  
4. Return to Kitchen Shelf and place box on middle position
5. Go to Kitchen Table and grab mug
6. Return to Kitchen Shelf and place mug on bottom position
Let me start by grabbing the cube since I'm already at the Front Door."
---
# Example of a tool call and response
[TOOL] Robot is trying to grab item cube
[TOOL-LOG] Grab activity initiated
[TOOL-LOG] Removed object cube from floor
[TOOL-RESPONSE] Successfully grabbed object
status: "success"
message: "Object grabbed successfully"
---
# The remaining tool calls during the execution are truncated for brevity
---
[TOOL] Robot looked around... 
---
[TOOL] The agent wrote to the scratchpad: "Perfect! Task completed successfully..."
---
[TOOL] Agent has ended task with result:
status: "success"
description: "Put away all the items into the shelf."
summary: "I successfully completed the task of putting away all items into the shelf. I systematically identified and organized three moveable items..."
---
[SYSTEM] Test results for task: success
\end{lstlisting}

    \section{EVALUATION}
    \label{sec:evaluation}

    To evaluate our proposed methodology for LLM-driven cognitive processing, we perform a series of experiments that investigate the system's ability to generate plans, execute them, and recover from failures, as well as the effect of episodic memory on the planning and execution success.
    In addition, we investigate how the performance of the system is affected by the use of different LLMs, using both open-weight and closed-source models.
    All evaluation is performed in the simulated 3D environment.

    \subsection{Evaluation Tasks}

    We evaluate the system on two tasks: placing items in a cupboard and swapping two items.
    \begin{description}
        \item[\textbf{T1: Placing items}] \hspace{1.75cm} In the first task, the agent has to place all items (a mug, a box, and a cube) into a cupboard with three shelves.
        For this, the agent needs to find the items, successfully navigate to each of them, grab them, then navigate to the cupboard, and place them in an unoccupied shelf.
        \item[\textbf{T2: Swapping items}] \hspace{2.1cm} The objective of the second task is to swap the mug and the cube.
        To perform the task, the agent needs to reason that it can only manipulate a single item at a time and that it thus needs to find a temporary storing location for one of the objects.
        In our evaluation setup, this task is only performed after task T1 is completed successfully, both for simplicity and for investigating long-horizon model performance.
    \end{description}
    For the purposes of fair quantitative evaluation, the task setup was static in all evaluation trials, namely the initial locations of the robot and objects were always the same.

    \subsection{Investigated LLMs}

    The evaluated models include GPT-4.1 \cite{gpt41}, Claude 4 Sonnet \cite{claude4}, Qwen3 Coder 480B A35B Instruct \cite{qwen3}, and DeepSeek V3.1 \cite{deepseek31}.
    GPT-4.1 and Claude 4 Sonnet are used through their respective official APIs, while quantised versions of Qwen3 Coder 480B A35B Instruct and DeepSeek V3.1 are hosted on a local cluster for inference. 
    The Qwen3 Coder 480B A35B Instruct model was hosted using vLLM \cite{kwon2023} with 4-bit Activation-Aware Quantization (AWQ) \cite{lin2024}, while the DeepSeek V3.1 was hosted using llama.cpp \cite{llamacpp} with 2-bit Unsloth Dynamic 2.0 GGUF \cite{unsloth} quantization. 
    It should be noted that quantization can degrade model quality and performance; however, this provides insight into whether on-premise hosted models can be used as alternatives to models hosted in the cloud.

    \subsection{Execution Success Evaluation}

    To evaluate the ability of our system to correctly solve the tasks, we repeat both tasks multiple times and verify the number of times the execution was successful; indirectly, this also indicates how often the robot was able to create a correct task plan.
    We consider the robot to have completed a task successfully if the state of the world (obtained from the simulation engine) meets the task requirements after the agent calls the \texttt{End-Task} tool, which we consider to be the ground-truth success (as opposed to the LLM's believed success). This also means that successful attempts include cases in which the agent had to recover during execution.
    On the other hand, a failure is considered if (i) the world state is not as expected at the end of the execution or (ii) the agent refuses to perform a task after $10$ repeated invocations.\footnote{Refusal occurs due to a hallucination when the model believes that it has already completed a task before it even starts performing it, and hence does not attempt the task.}

    The results of this evaluation are shown in Tab. \ref{tab:execution-success-t1} and Tab. \ref{tab:execution-success-t2}.
    \begin{table}[t]
        \caption{Success rate of the different LLMs on task T1}
        \label{tab:execution-success-t1}
        \begin{tabular}{L{0.18\linewidth} | M{0.14\linewidth} | M{0.14\linewidth} | M{0.14\linewidth} | M{0.14\linewidth}}
            & \cellcolor{gray!10}\textbf{GPT} & \cellcolor{gray!10}\textbf{Claude} & \cellcolor{gray!10}\textbf{Qwen} & \cellcolor{gray!10}\textbf{DeepSeek} \\\hline
            \cellcolor{gray!10}\textbf{\# executions} &      84 &   18 &     400 & 311 \\\hline
            \cellcolor{gray!10}\textbf{Success rate}  & $100\%$ & $100\%$ & $80\%$ & $100\%$
        \end{tabular}
    \end{table}
    \begin{table}[t]
        \caption{Success rate of the different LLMs on task T2}
        \label{tab:execution-success-t2}
        \begin{tabular}{L{0.18\linewidth} | M{0.14\linewidth} | M{0.14\linewidth} | M{0.14\linewidth} | M{0.14\linewidth}}
            & \cellcolor{gray!10}\textbf{GPT} & \cellcolor{gray!10}\textbf{Claude} & \cellcolor{gray!10}\textbf{Qwen} & \cellcolor{gray!10}\textbf{DeepSeek} \\\hline
            \cellcolor{gray!10}\textbf{\# executions} &      81 &         16 &    320 & 306 \\\hline
            \cellcolor{gray!10}\textbf{Success rate}         & $44.4\%$ & $100\%$ & $66.2\%$ & $75.5\%$
        \end{tabular}
    \end{table}
    In the tables, it can be seen that all models are able to reliably solve task T1, but they have varying levels of success on the more complex second task.
    Additionally, it can be seen that the different LLMs were executed a varied number of times, which is due to the fact that LLMs that are called through an API are considerably slower than the locally hosted counterparts.
    In particular, the results show that Claude 4 Sonnet has the highest success rate ($100\%$) across both tasks; however, during the trial period, we observed high response times of between $30$--$150s$ per request, which makes the model impractical to use on a robot platform that needs quick reactivity.

\begin{figure}[t]
        \begin{subfigure}[t]{0.475\linewidth}
            \centering
            \includegraphics[width=\linewidth]{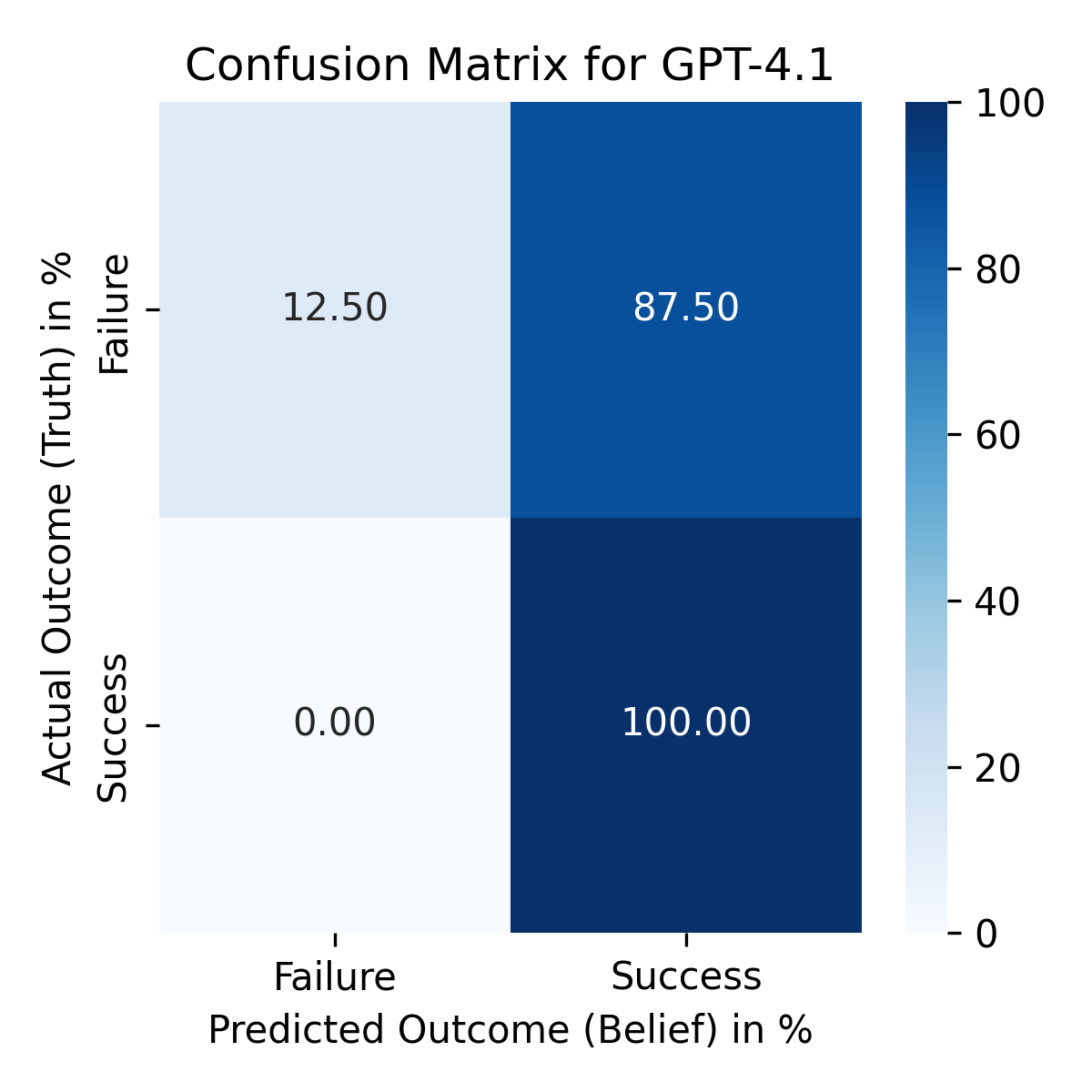}
            \caption{GPT-4.1}
            \label{fig:gpt-confusion-matrix}
        \end{subfigure}
        \hspace{0.025\linewidth}
        \begin{subfigure}[t]{0.475\linewidth}
            \centering
            \includegraphics[width=\linewidth]{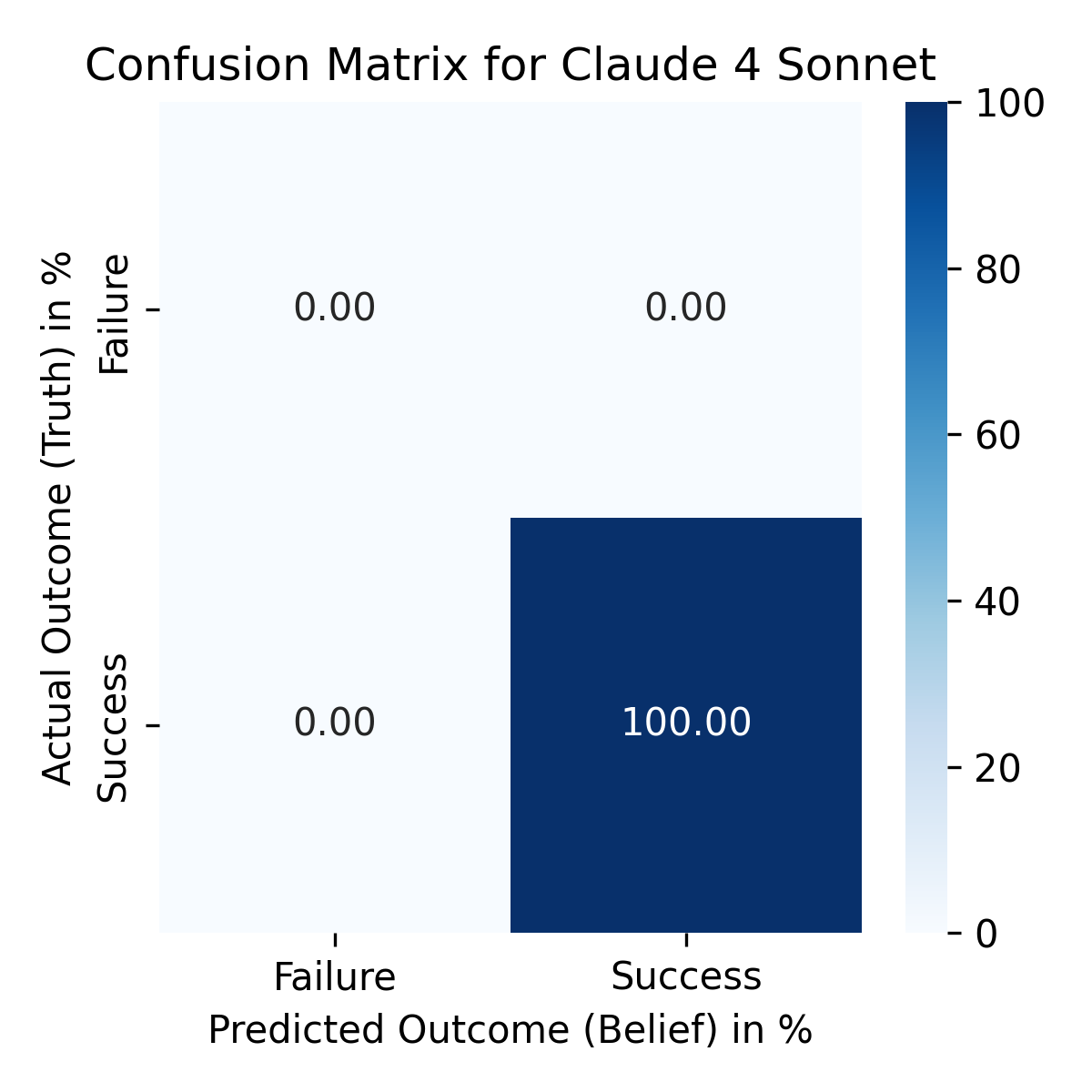}
            \caption{Claude 4 Sonnet}
            \label{fig:claude-confusion-matrix}
        \end{subfigure}
        \newline
        \begin{subfigure}[t]{0.475\linewidth}
            \centering
            \includegraphics[width=\linewidth]{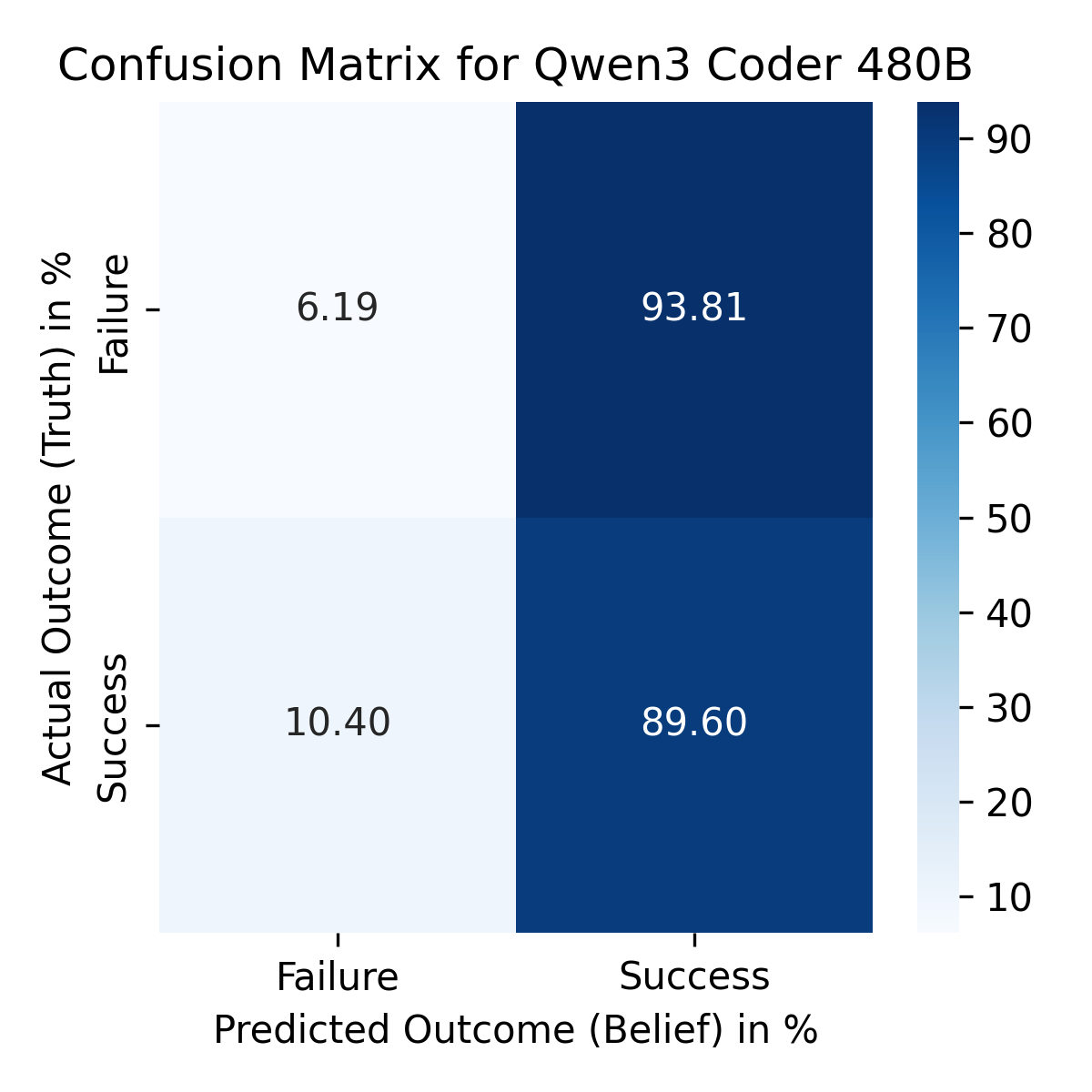}
            \caption{Qwen3 Coder 480B}
            \label{fig:qwen-confusion-matrix}
        \end{subfigure}
        \hspace{0.025\linewidth}
        \begin{subfigure}[t]{0.475\linewidth}
            \centering
            \includegraphics[width=\linewidth]{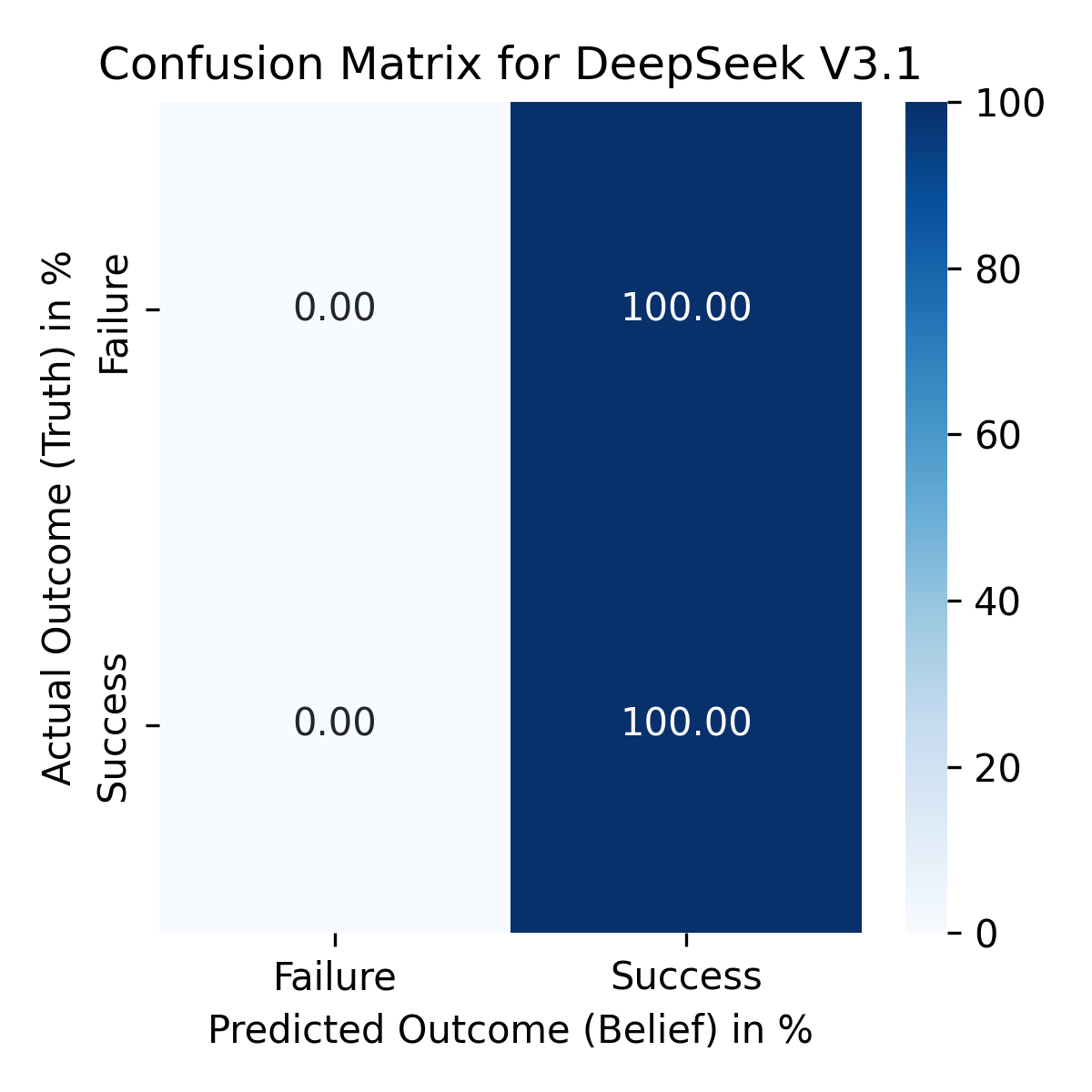}
            \caption{DeepSeek V3.1}
            \label{fig:deepseek-confusion-matrix}
        \end{subfigure}
        \caption{Confusion matrices illustrating the models' believed execution success as opposed to the actual execution success}
        \label{fig:confusion-matrices}
    \end{figure}

    The results displayed in Tab. \ref{tab:execution-success-t1} and \ref{tab:execution-success-t2} are the actual execution success in the evaluation trials; however, the models can also hallucinate success even when they fail.
    This discrepancy between the actual execution success and the believed execution success has significant effects on the real-world usability of LLMs for autonomous task planning and execution.
    In Fig. \ref{fig:confusion-matrices}, we show confusion matrices that illustrate these metrics in the evaluation trials.
    As can be seen, all models except Claude (which did not fail in any of the trials) hallucinate success; DeepSeek V3.1 is the only one that is consistently overconfident about its success, namely it believes that it has succeeded in all execution trials, but they all tend to significantly overestimate their performance.

    \subsection{Benefits of Memory on Planning}

\begin{figure}[t]
        \begin{subfigure}[t]{0.475\linewidth}
            \centering
            \includegraphics[width=\linewidth]{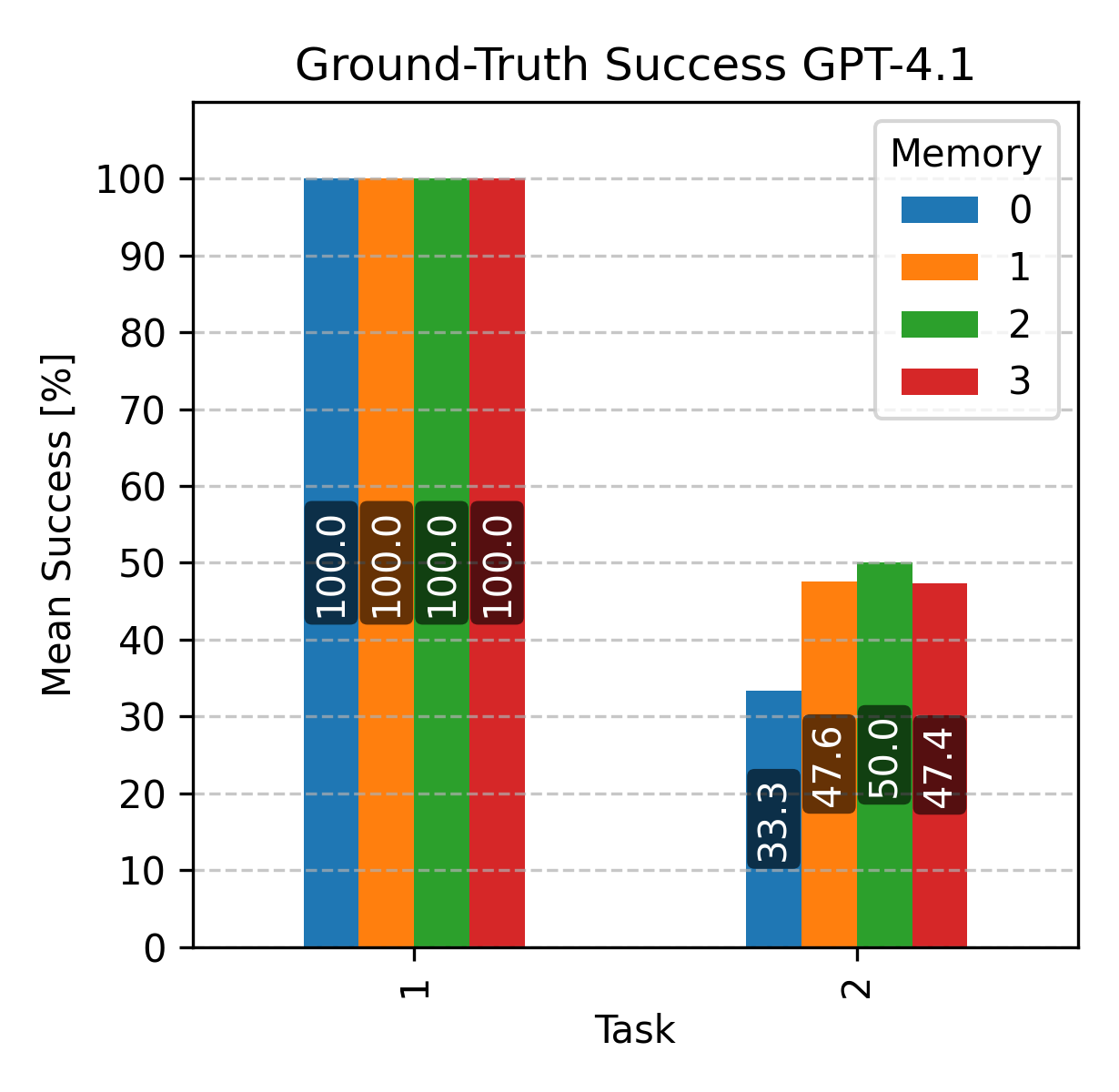}
            \caption{GPT-4.1}
            \label{fig:gpt-memory-accuracy}
        \end{subfigure}
        \hspace{0.025\linewidth}
        \begin{subfigure}[t]{0.475\linewidth}
            \centering
            \includegraphics[width=\linewidth]{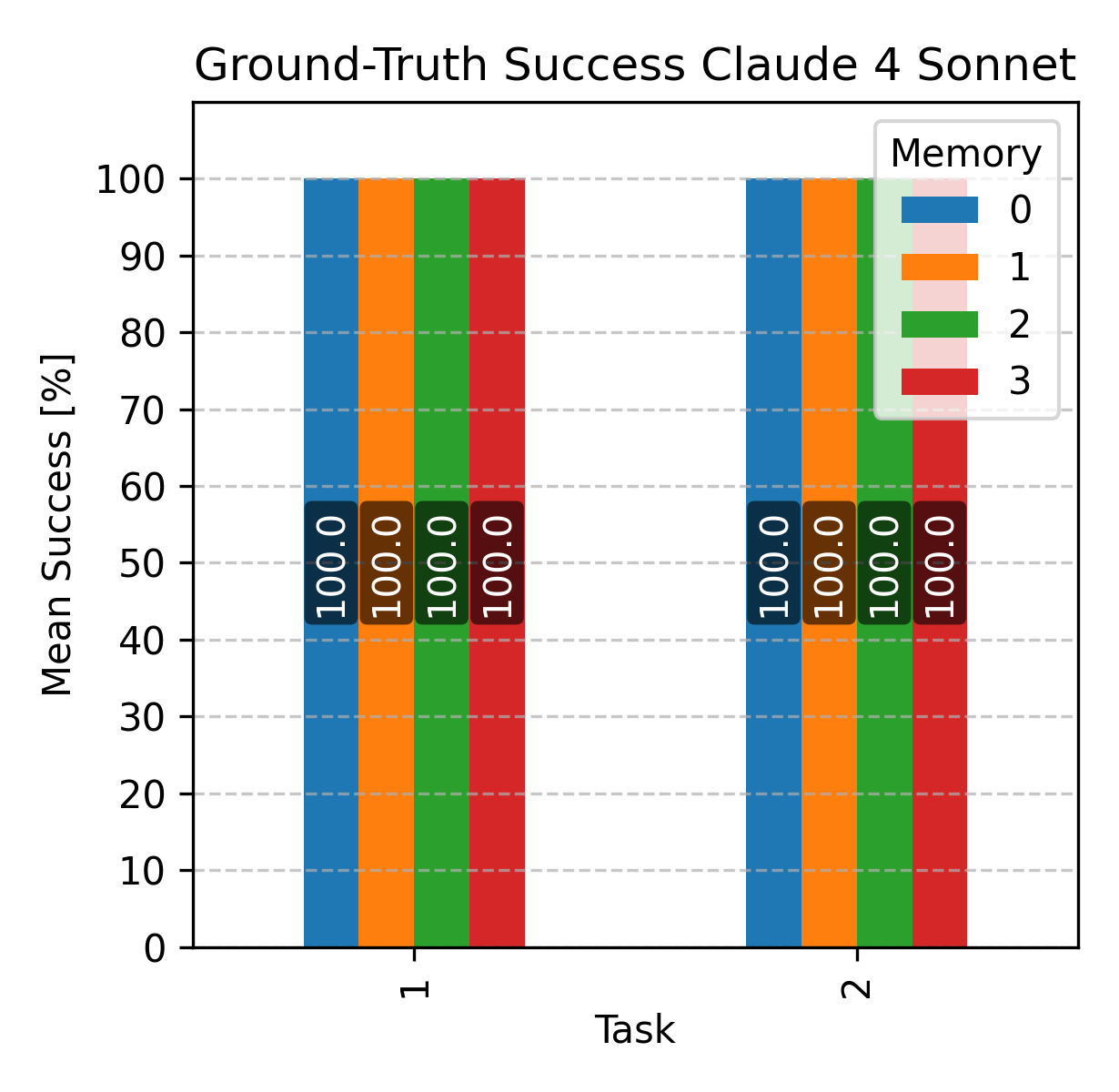}
            \caption{Claude 4 Sonnet}
            \label{fig:claude-memory-accuracy}
        \end{subfigure}
        \newline
        \begin{subfigure}[t]{0.475\linewidth}
            \centering
            \includegraphics[width=\linewidth]{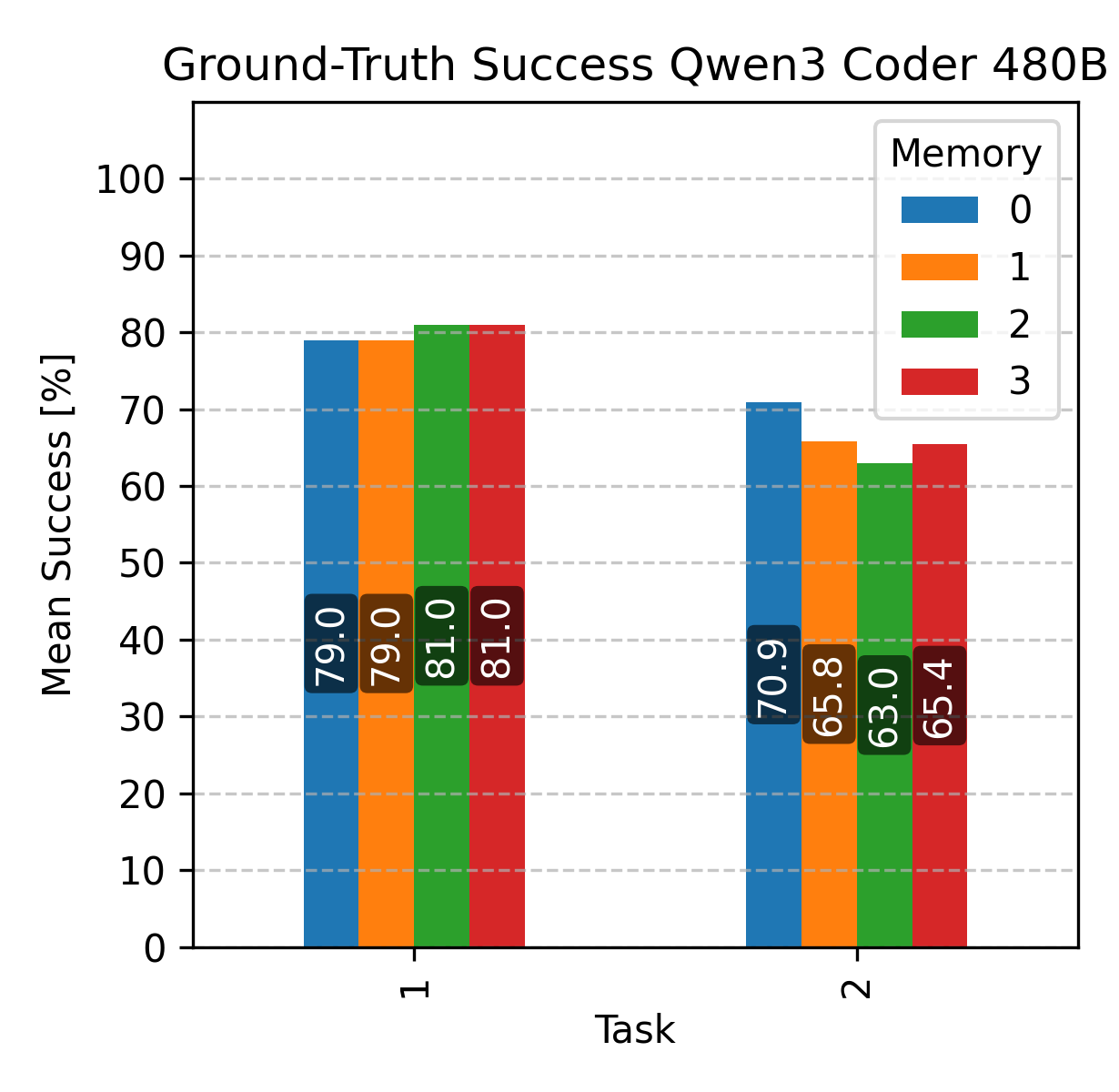}
            \caption{Qwen3 Coder 480B}
            \label{fig:qwen-memory-accuracy}
        \end{subfigure}
        \hspace{0.025\linewidth}
        \begin{subfigure}[t]{0.475\linewidth}
            \centering
            \includegraphics[width=\linewidth]{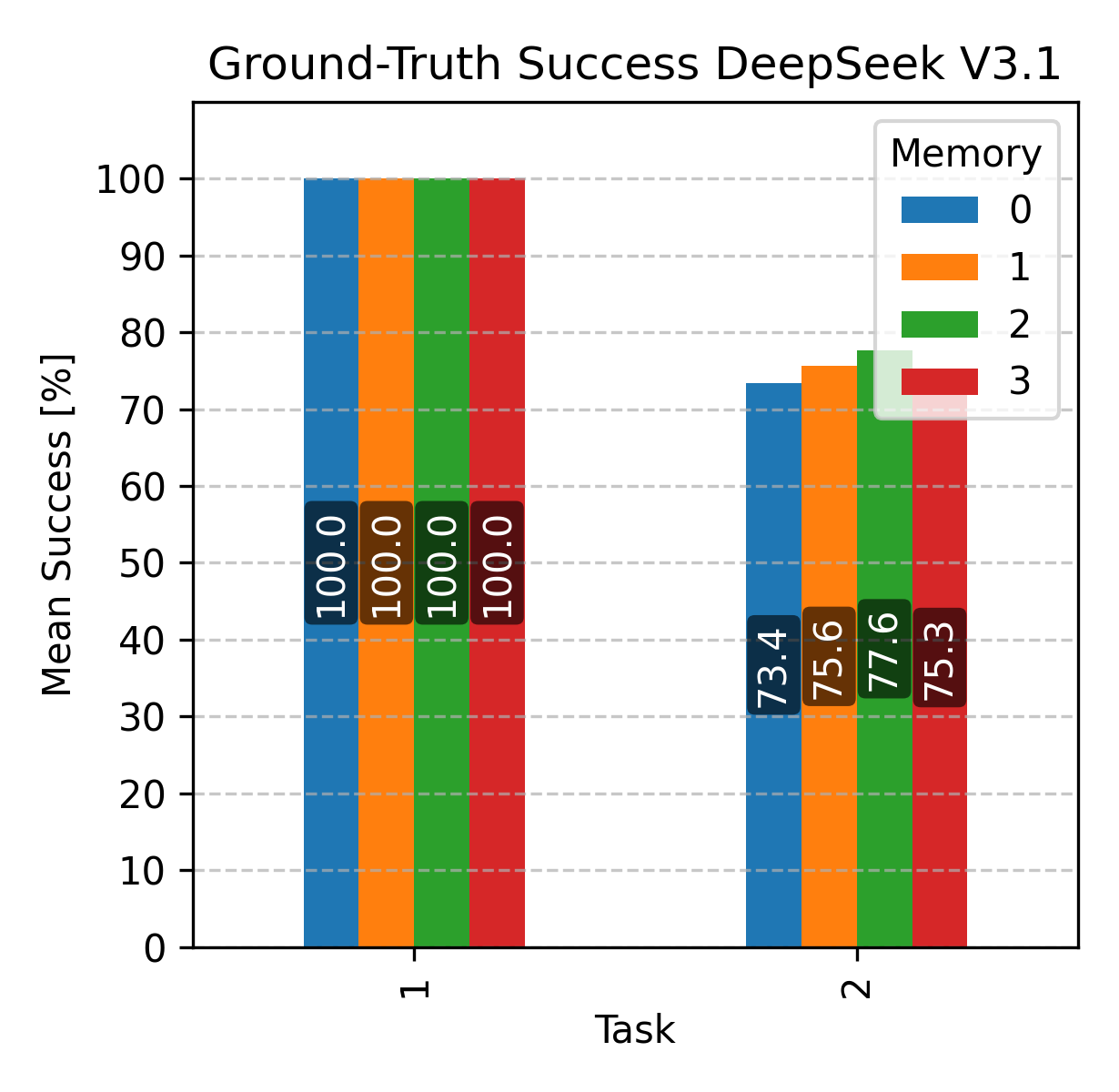}
            \caption{DeepSeek V3.1}
            \label{fig:deepseek-memory-accuracy}
        \end{subfigure}
        \caption{Ground-truth success rate of the models over an increasing number of executions added to the episodic memory}
        \label{fig:memory-accuracy}
    \end{figure}

    In addition to evaluating the overall success rate, we also evaluate the effect of adding episodic memories on the execution success, as this shows whether the LLM-based agent can benefit from prior executions when reasoning about plans.
    For this evaluation, we compare (a) the ground-truth success rate of the models before any executions are included in the episodic memory with (b) the success rate as we include memories (up to three executions).
    The results are shown in Fig. \ref{fig:memory-accuracy}.\footnote{It should be noted that the executions for this evaluation are the same ones that are used to compute the overall success rate in Tab. \ref{tab:execution-success-t1} and \ref{tab:execution-success-t2}; in other words, the tables present an aggregated success rate.}

    Based on these results, the overall effect of the memory on the task completion success is inconclusive, and varies between models and tasks.
    For GPT-4.1 and DeepSeek V3.1, we can notice a small improvement of the execution success as more memories are accumulated; however, for Qwen3 Coder, there is a small improvement for task T1, but a small decrease in performance for task T2.
    We believe that this is due to the possibility of adding incorrectly labelled executions to the episodic memory; this is possible because memories are labelled with the agent's belief about its execution success, and this belief can be hallucinated.

\begin{figure}[t]
        \begin{subfigure}[t]{0.475\linewidth}
            \centering
            \includegraphics[width=\linewidth]{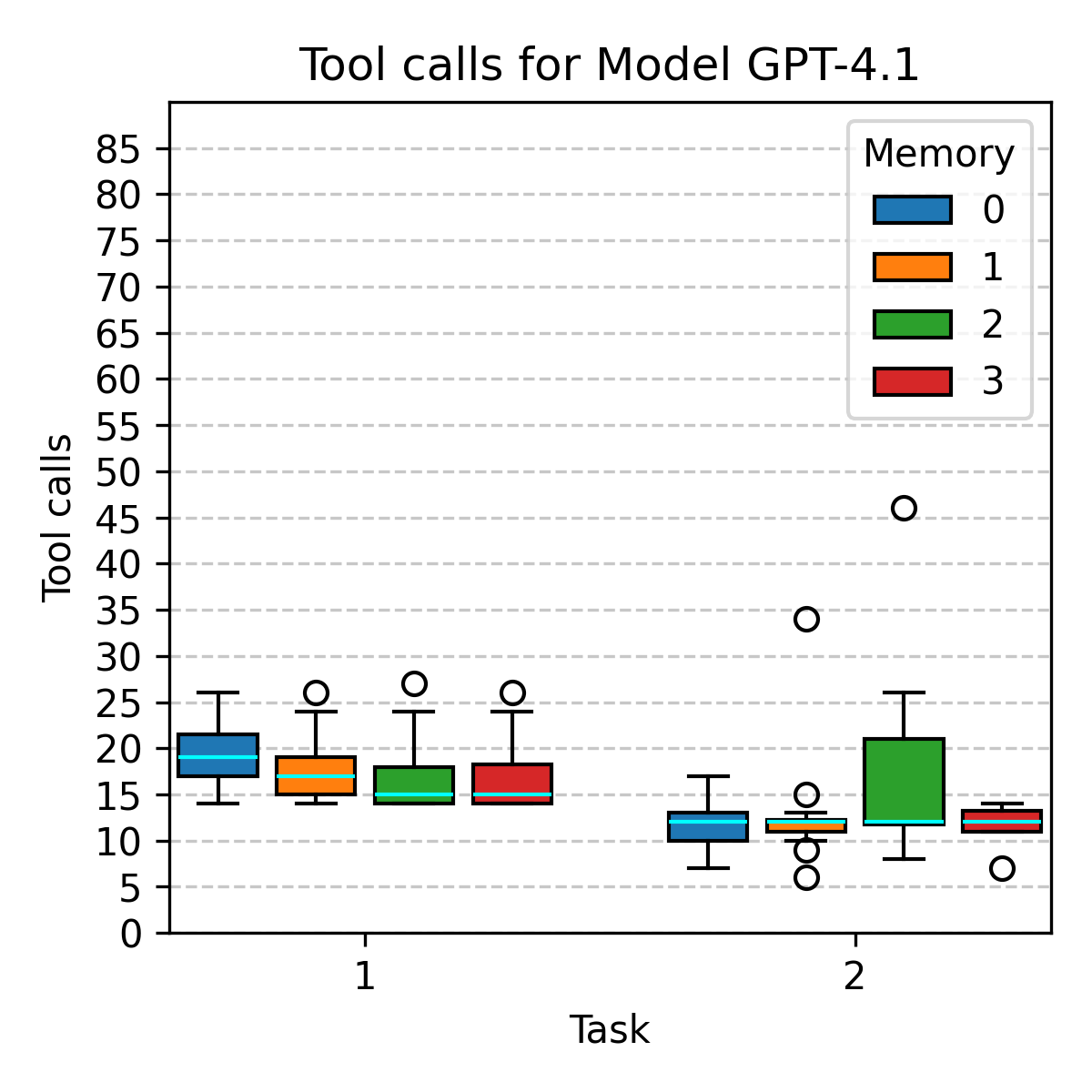}
            \caption{GPT-4.1}
            \label{fig:gpt-tool-calls}
        \end{subfigure}
        \hspace{0.025\linewidth}
        \begin{subfigure}[t]{0.475\linewidth}
            \centering
            \includegraphics[width=\linewidth]{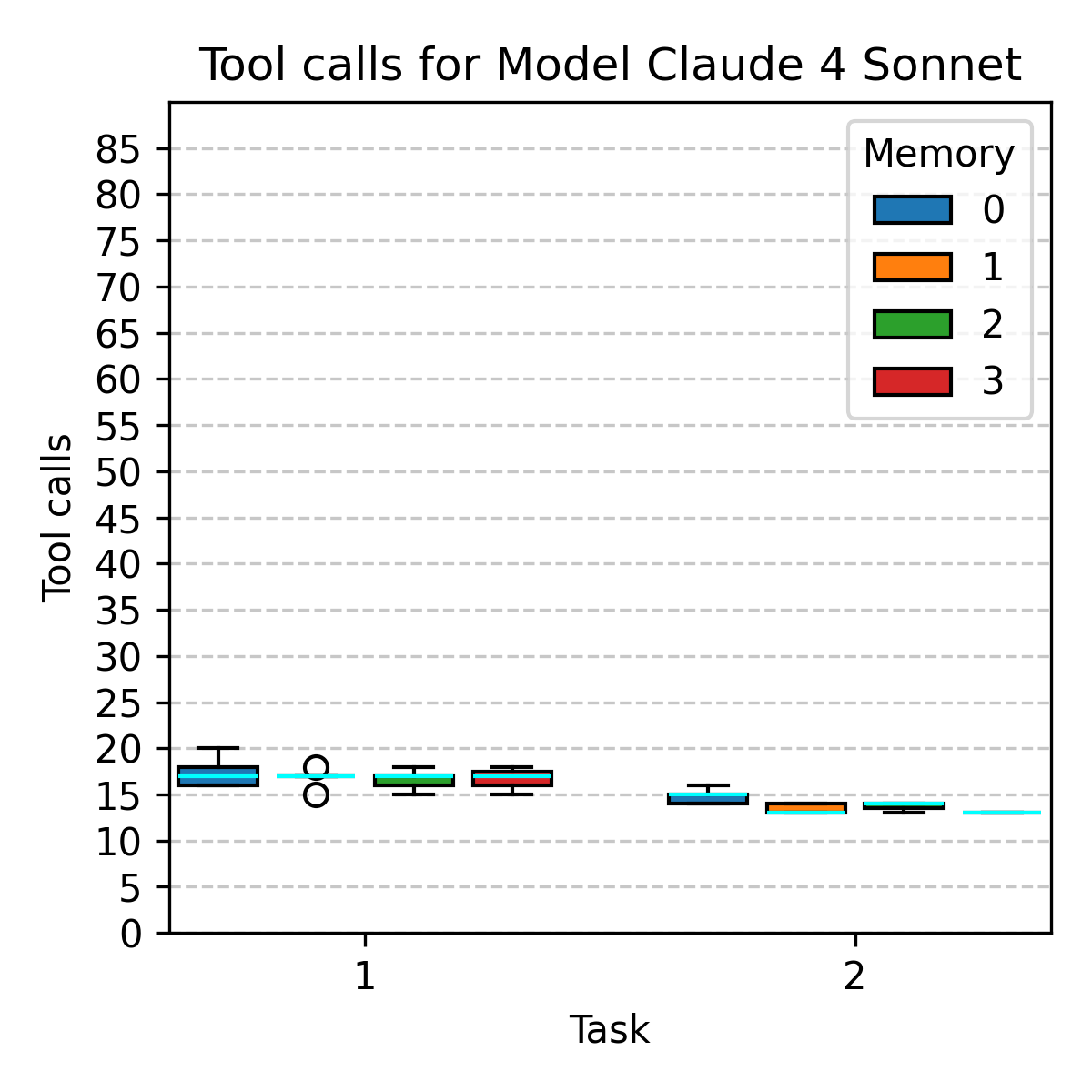}
            \caption{Claude 4 Sonnet}
            \label{fig:claude-tool-calls}
        \end{subfigure}
        \newline
        \begin{subfigure}[t]{0.475\linewidth}
            \centering
            \includegraphics[width=\linewidth]{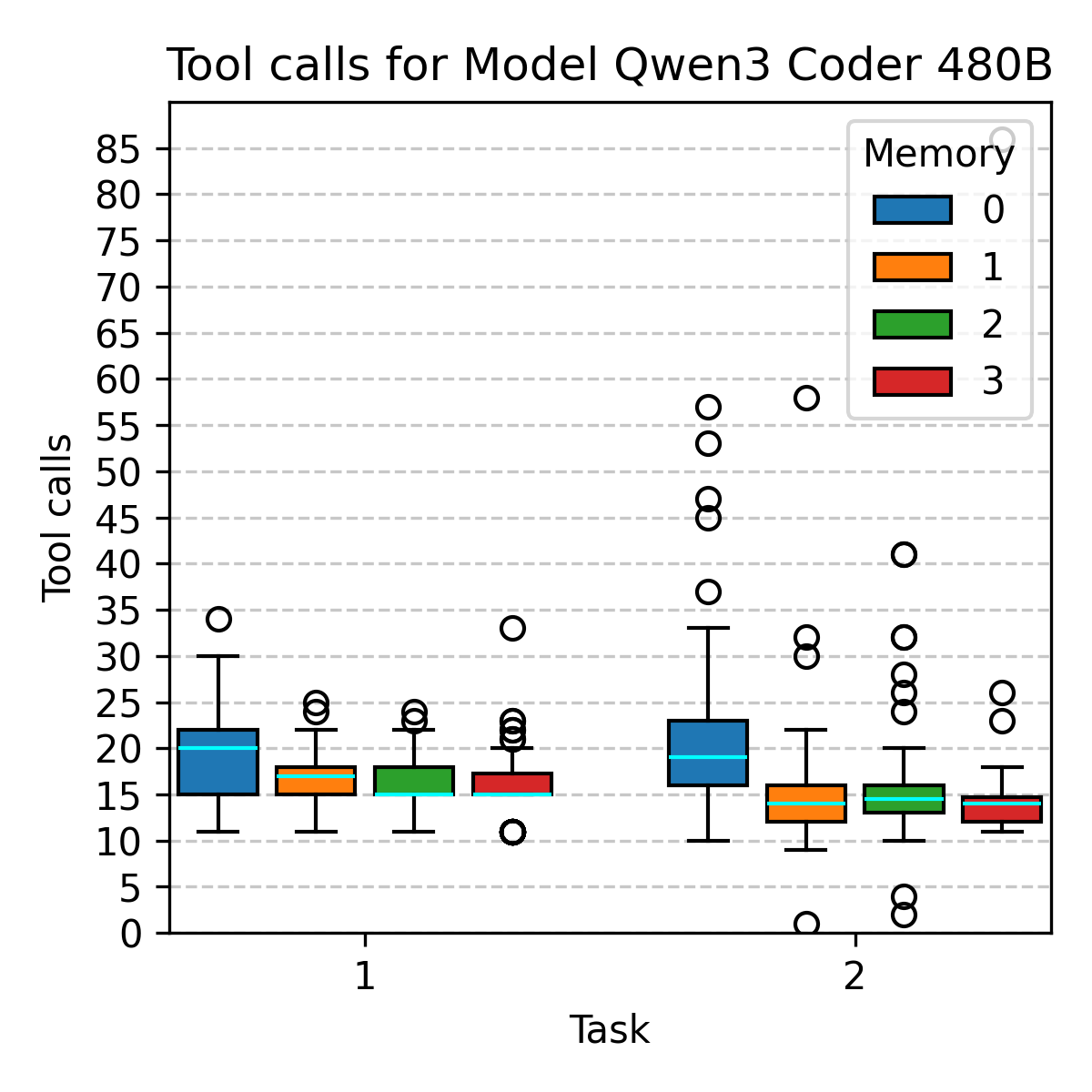}
            \caption{Qwen3 Coder 480B}
            \label{fig:qwen-tool-calls}
        \end{subfigure}
        \hspace{0.025\linewidth}
        \begin{subfigure}[t]{0.475\linewidth}
            \centering
            \includegraphics[width=\linewidth]{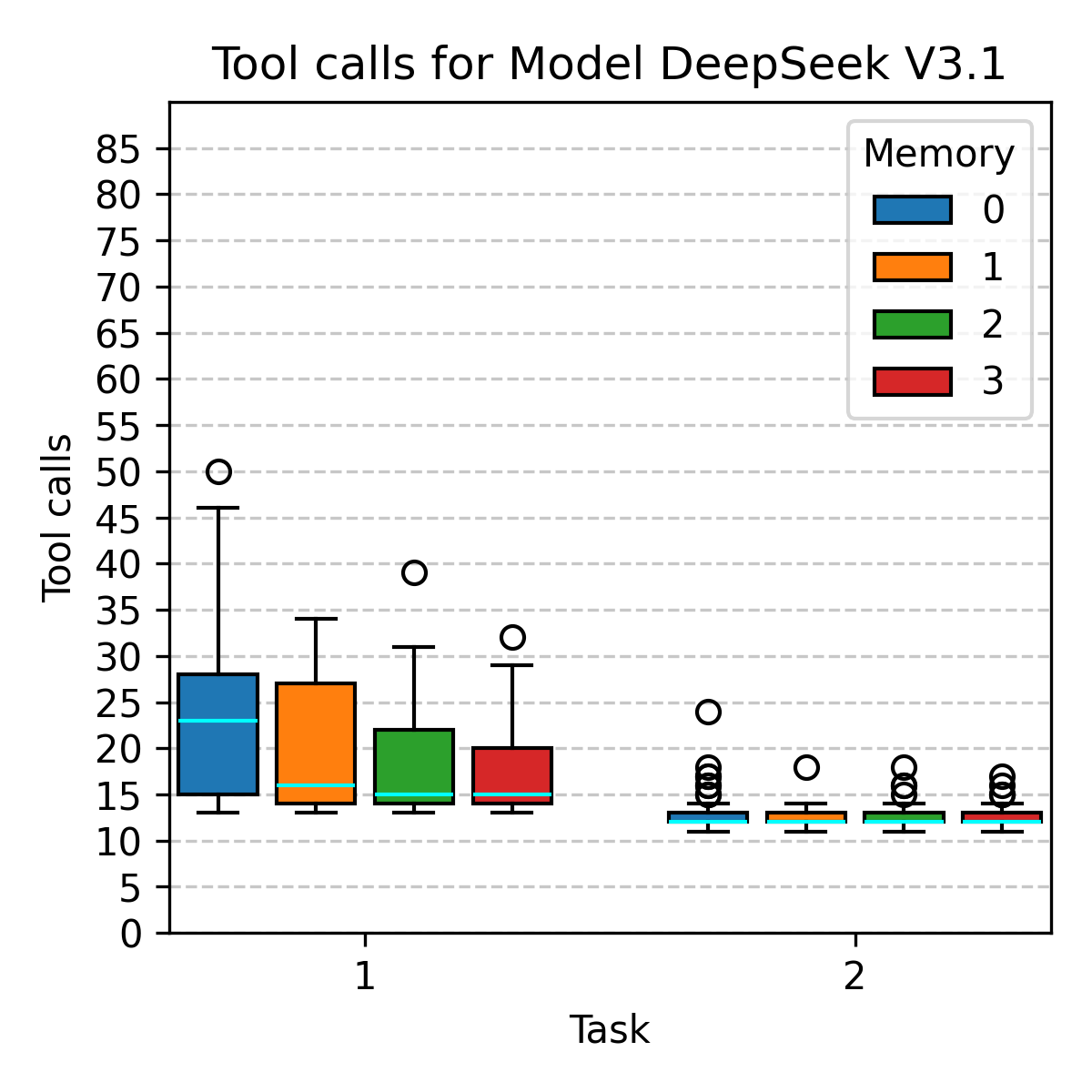}
            \caption{DeepSeek V3.1}
            \label{fig:deepseek-tool-calls}
        \end{subfigure}
        \caption{Tool calls of the models over an increasing number of executions accumulated in the episodic memory}
        \label{fig:tool-calls}
    \end{figure}

    In addition to evaluating the effect of the memory on the execution success, we compare the number of tool calls made by the different models as the memory size increases; this provides information about the potential benefits of the memories on avoiding unnecessary tool calls.
    Fig. \ref{fig:tool-calls} shows the results of this comparison.

    These results indicate that, on average, memories do indeed have a positive effect on reducing the number of tool calls, which is generally consistent for all tasks and models.

    \subsection{Qualitative Observations}

    Throughout the evaluation, we also made various qualitative observations that are important to discuss so that the results in this section can be appreciated better.

    One observation is that execution failures often occurred due to missing explicit knowledge about affordances in the environment.
    For instance, during the execution of task T1, many models would take the task prompt \texttt{Please put all the objects away into the shelf} quite literally and attempt to move a large TV object onto a shelf.
    The design of the environment prevents this, however, and agents would report a failure when attempting to grab the TV.
    Despite this misinterpretation, the intended objects (the mug, box, and cube) would indeed be correctly moved into the cupboard by all models that were evaluated.
    This shows that, while the agents were capable of performing the core task, they sometimes struggled with interpreting ambiguous instructions; cases like this are where the memory is particularly useful, as the robot can avoid repeating actions that are known to fail.

    Related to this, we also observed that the models have an inherent ability to recover from failed actions, such as placing in occupied spaces or attempting to pick an object while already holding something else; in such cases, the models tend to automatically replan without any human intervention.
    This recovery is, however, sometimes affected by hallucinations, which can then lead to other failures, even though the model believes that it has successfully recovered.

    With respect to the evaluated models, it should be mentioned that the ones used are models that we observed to be able to coherently perform agentic reasoning (after a qualitative assessment over sample executions).
    In our pre-testing, we also experimented with other models, but we could not achieve meaningful results with them.
    Concretely, we noticed that smaller models (around $24B$ parameters) were unable to coherently chain together actions that would enable the robot to make progress on the experimental tasks, or would get stuck in an infinite generation loop.

    \section{DISCUSSION}
    \label{sec:discussion}

    This paper investigated whether an LLM can serve as the cognitive core of an embodied robotic agent in a simulated household environment.
    By embedding an LLM within a cognitive architecture that combines working memory, episodic memory, and a tool-calling interface, we enabled a mobile manipulator to perceive, plan, and act in a physically grounded setting.
    The robot successfully executed object manipulation and navigation tasks, such as placing items in a cupboard and reordering them across locations.
    Our evaluation shows that LLMs are generally capable of reasoning about spatial layouts, sequencing tool calls, and adapting strategies when faced with environmental constraints; additionally, the implementation of episodic memory proved somewhat valuable, as agents slightly improved in task completion when they could draw upon past experiences.
    However, there are also notable limitations that need to be addressed for robust real-world deployment; this includes overconfidence (which leads to an incorrect belief about the actual task success and hampers the effectiveness of the episodic memory), refusal to perform tasks (particularly after completing a prior task), inability to efficiently recover from unsuccessful tool calls and plans (requiring several unsuccessful attempts before successfully replanning), and occasional misinterpretation of ambiguous instructions.

    Future work should focus on improving the task generalisation and robustness by coupling the procedure of adding experiences to the episodic memory with an external fact-checking component that verifies the LLM's belief about task success; this could be a human in the loop, but could also be a dedicated model to evaluate the robot's actual task success.
    Another important aspect would be the integration of an automated evaluation pipeline, which could generate quantitative performance benchmarks across a wider variety of tasks, models, and robot platforms, thereby enabling systematic comparison of different architectural approaches and LLM capabilities.
    With respect to our simulation, future work will focus on extending it towards richer environments with more complex objects, dynamic obstacles, and potentially multi-agent interactions to enable a closer approximation of real-world deployment scenarios.
    As the focus of this work was to evaluate the overall ability of LLM-based systems to be at the core of a cognitive robot architecture, we performed simulation-based evaluation to abstract away challenges of real-world systems, such as sensor noise, mechanical wear, and latency constraints; in future work, we intend to implement the architecture on a physical robot to validate the extensibility of our approach.
    Finally, it would be worthwhile to perform a systematic comparison of our LLM-based architecture with other cognitive architectures; this could provide insights on the strengths and weaknesses of various aspects included in different architectures.


\addtolength{\textheight}{-4cm}   

\bibliographystyle{IEEEtran}
\bibliography{references}

\end{document}